\documentclass{article}
\usepackage{graphicx}      
\usepackage[authoryear]{natbib}
\setcitestyle{authoryear}
\usepackage{amsmath} 
\usepackage{amssymb}  
\usepackage{url}
\usepackage{authblk}
\usepackage{subcaption}

\usepackage{algorithm}

\usepackage{enumitem}
\renewcommand{\theenumi}{\arabic{enumi}}

\renewcommand{\theenumii}{\arabic{enumii}}

\newtheorem{remark}{Remark}{\normalfont}{\normalfont}
\newcommand{\thetao}{{\theta_{\rm o}}}
\newcommand{\So}{{S}}            
\newcommand{\M}{{M}}             
\newcommand{\MS}{\mathcal{M}}    
\newcommand{\D}{{\mathcal{D}}}
\newcommand{\NN}{\mathcal{N\!N}} 
\newcommand{\batchsize}{q}
\newcommand{\seqlen}{m}
\newcommand{\nin}{n_u} 
\newcommand{\ny}{n_y} 
\newcommand{\nx}{n_x}
\newcommand{\numiter}{n}
\newcommand{\nsamp}{N}
\newcommand{\tens}[1]{\mathbf{#1}}
\newcommand{\myeq}[1]{\mathrel{\overset{\makebox[0pt]{\mbox{\normalfont\tiny\sffamily $#1$}}}{=}}}
\newcommand{\ntheta}{n_\theta}
\newcommand{\Yid}{Y}
\newcommand{\Uid}{U}
\newcommand{\Did}{{\mathcal{D}}}
\newcommand{\hidden}[1]{\overline{#1}}
\newcommand{\ymodel}{y^{\rm o}} 
 
\newcommand{\norm}[1]{\left\lVert#1\right\rVert}

\title{Model structures and fitting criteria for system identification with neural networks} 

\author[1]{Marco Forgione}
\author[1]{Dario Piga}
\affil[1]{IDSIA Dalle Molle Institute for Artificial Intelligence SUPSI-USI, Manno, Switzerland}  

\begin{document}
\maketitle

\vskip 1em
\noindent\rule{\textwidth}{1pt}
 To cite this work, please use the following bibtex entry:
\begin{verbatim}
@inproceedings{forgione2020model,
  title={Model structures and fitting criteria for system identification
  with neural networks},
  author={Forgione, Marco and Piga, Dario},
  booktitle={Proc. of the 14th IEEE International Conference Application
  of Information and Communication Technologies, Tashkent, Uzbekistan},
  year={2020}
}
\end{verbatim}
\vskip 1em
Using the plain bibtex style, the bibliographic entry should look like:\\ \\
M. Forgione, D. Piga. Model structures and fitting criteria for system identification with neural networks. In \textit{Proc. of the 14th IEEE International Conference Application of Information and Communication Technologies, Tashkent, Uzbekistan}, 2020.

\noindent\rule{\textwidth}{1pt}

\begin{abstract}                
This paper focuses on the identification of dynamical systems with tailor-made model structures, where neural networks are used to approximate uncertain components and domain knowledge is retained, if  available. These model structures are fitted  to measured data using different criteria
 including a computationally efficient approach minimizing a regularized multi-step ahead simulation error. The neural network parameters are  estimated along with the initial conditions used to simulate the output signal in small-size subsequences. A regularization term is included in the fitting cost in order to enforce these initial conditions to be consistent with the estimated system dynamics.\footnote{This work was partially supported by the European H2020-CS2 project ADMITTED, Grant agreement no. GA832003.}
\end{abstract}


\section{Introduction}
In recent years, deep learning has advanced at a tremendous pace and  is now the core methodology behind cutting-edge technologies such as speech recognition, image classification and captioning, language translation, and autonomous driving \citep{schmidhuber2015deep}. These impressive achievements are attracting ever increasing investments both from the private and the public sector, fueling further research in this field. 

A good deal of the advancement in the deep learning area is of public domain, both in terms of scientific publications and software tools.
 Nowadays, highly optimized and user-friendly deep learning frameworks are available \citep{paszke:2017automatic}, often distributed under permissive open-source licenses.
Using the high-level functionalities of a deep learning framework and following good practice, even a novice user can deal with  \emph{standard}  machine learning tasks (once considered extremely hard) such as image classification with moderate effort. Under the hood, the
machine learning task is automatically transformed into a relevant optimization problem and subsequently solved through efficient numerical routines.

 An experienced practitioner can employ the same deep learning framework at a lower level to tackle non-standard learning problems, by defining customized models and objective functions to be optimized, and using operators such as \emph{neural networks} as building blocks. 
The practitioner is free from the burden of writing optimization code from scratch for every particular problem, which would be tedious and error-prone.
In fact, as a built-in feature, modern deep learning engines can compute the derivatives of a supplied objective function with respect to free tunable parameters by implementing the celebrated \emph{back-propagation} algorithm \citep{rumelhart1988learning}. In turn, this enables convenient setup of any gradient-based optimization method.

An exciting, challenging---and yet largely unexplored---application field is \emph{system identification} with tailor-made model structures and fitting criteria. In this context, neural networks can be used to describe uncertain components of the  dynamics, while retaining structural (physical) knowledge, if available. Furthermore, the fitting criterion can be specialized to take into account the modeler's ultimate goal, which could be prediction, failure detection, state estimation, control design, simulation, \emph{etc}.  

The choice of the cost function may also be influenced by computational considerations. 
In this paper, in particular, models are evaluated according to their simulation performance. In this setting, from a theoretical perspective, simulation error minimization is generally the best fitting criterion. However, computing the simulation error loss and its derivatives may be prohibitively expensive from a computational perspective for   dynamical models involving neural networks.  
We  show that multi-step simulation error minimization over \emph{batches} of small-size subsequences extracted from the identification dataset provides models with high simulation performance, while keeping the computational burden of the fitting procedure acceptable. In the proposed method, the neural network parameters are jointly  estimated  with  the  initial conditions used to simulate the system in each subsequence.  
A regularization term is also included in the fitting criterion in order to enforce all these initial conditions to be consistent with the estimated system dynamics.

The use of neural networks in system identification has a long history, see, \emph{e.g.}, \citep{werbos1989neural, chen1990non}. Even though motivated by similar reasoning, these earlier works are hardly comparable given the huge gap of hardware/software technology.    
More recently, a few interesting approaches using modern deep learning tools and concepts in system identification have been presented. For instance,   \citep{masti2018learning} introduces a technique to identify neural state-space model structures using deep autoencoders for state reconstruction, while 
\citep{gonzalez2018non, wang2017new} discuss the use of \emph{Long Short-Term Memory} (LSTM) recurrent neural networks for system identification.
Compared to these recent contributions, our work focuses on using specialized model structures for the identification task at hand. In the machine learning community, neural networks have also been recently applied for approximating the solution of ordinary and partial differential equation, see \emph{e.g.}, \citep{chen2018neural, raissi2019physics}. With respect to these contributions, our aim  is to find computationally efficient fitting strategies that are robust to the measurement noise.

The  rest of this paper is structured as follows. The overall settings and problem statement is outlined in Section \ref{sec:settings}. The neural dynamical model structures are introduced in Section \ref{sec:model_structure} and   criteria for fitting these model structures to training data are described in Section \ref{sec:training}. Simulation results are presented in Section \ref{sec:example} and can be replicated using the codes available at \url{https://github.com/forgi86/sysid-neural-structures-fitting}.  Conclusions and directions  for future research are  discussed in Section \ref{sec:conclusions}. 

\section{Problem Setting}
\label{sec:settings}
We are given a dataset $\Did$ consisting of $\nsamp$ input samples $\Uid = \{u_{0},\;u_{1},\dots,\;u_{\nsamp-1}\}$ and output samples  $\Yid = \{y_{0},\;y_{1},\dots,\;y_{\nsamp-1}\}$, gathered from an experiment on a  dynamical system $\So$.  The data-generating system $\So$ is assumed to have the discrete-time  state-space representation
\begin{subequations}
\label{eq:data_generating}
\begin{align}
 x_{k+1} &= f(x_{k}, u_{k}) \\
 y^{\text o}_k     &= g(x_k),
\end{align}
\end{subequations}
where $x_k \in \mathbb{R}^{\nx}$ is the state at time $k$; $y^{\text o}_k  \in \mathbb{R}^{\ny}$ is the noise-free output; $u_k  \in \mathbb{R}^{\nin}$ is the input; $f(\cdot, \cdot)$ and $g(\cdot)$ are the state and output mappings, respectively.
The measured output $y_k$ is  corrupted by a zero-mean  noise $\eta_k$, \emph{i.e.}, $y_k =  y^{\text o}_k  + \eta_k$. 

The aim of the paper is twofold:
\begin{itemize}
	\item to introduce flexible neural model structures that are suitable to represent generic dynamical systems as~\eqref{eq:data_generating}, allowing the modeler to embed domain knowledge to various degrees and to exploit neural networks' universal approximation capabilities (see  \citep{hornik1989multilayer}) to describe unknown model components;
	\item to present  robust and computationally efficient procedures to fit these neural model structures to the training dataset  $\Did$.
\end{itemize}

\subsection{Full model structure hypothesis}
Let us consider a  \textit{model structure} $\MS = \{\M(\theta),\; \theta \in \mathbb{R}^{\ntheta}\}$, where $M(\theta)$ represents a dynamical model parametrized by a real-valued vector $\theta$.  We refer to \textit{neural model structures} as structures $\MS$ where some components of the model $M(\theta)$ are described by neural networks.


 Throughout the paper, we will make  the following \emph{full model structure} hypothesis:  there exists a parameter $\thetao \in \mathbb{R}^{\ntheta}$ such that the model $\M(\thetao) \in \MS$ is a perfect representation of the true system $\So$, \emph{i.e.},  for every  input sequence,  $\M(\thetao)$ and $\So$ provide the same output. We denote this condition as  $\M(\thetao) = \So$. Note that the parameter $\thetao$ may not be unique. Indeed, deep neural networks  have multiple equivalent representations obtained, for instance, by permuting neurons in a hidden layer. Let  $\Theta_\So$ be the set of parameters that provide a perfect system description, namely $\Theta_\So = \{\theta \in \Theta \text{ such that } \M(\theta) = \So\}$.  Under the full model structure hypothesis, the ultimate identification goal is to find a parameter $\theta \in \Theta_\So$. \\


\begin{remark}
In practice, fitting is performed on a finite-length dataset $\Did$ covering a finite number of experimental conditions. To this aim, let us introduce the notation $\M(\theta) \myeq{\D} \So$ meaning that 
the model $\M(\theta)$ perfectly matches $\So$ on the dataset $\D$ and    let  us define the parameter set  $\Theta_\So^\D = \{\theta \in \Theta \text{ such that } \M(\theta) \myeq{\D} \So\}$. By definition,   $\Theta_\So \subseteq \Theta_\So^\D$.  Thus, when fitting the model structure to a finite-length dataset $\Did$, we aim to find a parameter $\theta \in \Theta_\So^{\Did}$ (but not necessarily in $\Theta_{\So}$).
\end{remark}

\section{Neural model structures}
\label{sec:model_structure}
In this section, we introduce possible neural model structures for dynamical systems. 

\subsection{State-space structures}
\label{sec:ss_model_structure}
A general \emph{state-space neural  model structure} has form 
\begin{subequations}
\label{eq:neural_ss_general}
\begin{align}
 x_{k+1} &= \NN_x(x_{k}, u_{k}; \theta) \\
 \ymodel &= \NN_y(x_k; \theta),
\end{align}
\end{subequations}
where $\NN_x$ and $\NN_y$ are feedforward neural networks of compatible size parametrized by $\theta \in \mathbb{R}^{\ntheta}$.  Such a general structure can be tailored for the identification task at hand. Examples are reported in the following paragraphs.

\paragraph{Residual model}
If a linear approximation of the system is available, an appropriate model structure is
\begin{subequations}
\label{eq:neural_ss_lin}
\begin{align}
 x_{k+1} &= A_L x_k + B_L u_k + \NN_x(x_{k}, u_{k}; \theta),  \label{eq:neural_ss_lin_state}\\
 y_k     &= C_L x_k + \NN_y(x_{k}, u_{k}; \theta),
\end{align}
\end{subequations}
where $A_L$, $B_L$, and $C_L$ are matrices of compatible dimensions describing the linear system approximation. Even though model \eqref{eq:neural_ss_lin} is not more general than \eqref{eq:neural_ss_general}, it could be easier to train  as the neural networks $\NN_x$ and $\NN_y$ are supposed to capture only residual (nonlinear) dynamics.

\paragraph{Integral model}
When fitting data generated by a continuous-time system, the following neural model  with an integral term in the state equation can be used to encourage continuity of the solution:
\begin{subequations}
	\label{eq:neural_ss_full_Int}
	\begin{align}
	x_{k+1} &= x_k + \NN_x(x_{k}, u_{k}; \theta) \label{eq:neural_ss_state_int} \\ 
	\ymodel     &= \NN_y(x_{k}, u_{k}; \theta).
	\end{align}
\end{subequations}
 This structure can also be interpreted as the forward Euler discretization scheme applied to an underlying continuous-time  state-space model.

\paragraph{Fully-observed state model}
If the system state is known to be fully observed, an effective representation is
\begin{subequations}
\label{eq:neural_ss_full}
\begin{align}
 x_{k+1} &= \NN_x(x_{k}, u_{k}; \theta) \label{eq:neural_ss_state} \\ 
 \ymodel     &= x_k,
\end{align}
\end{subequations}
where only the state mapping neural network $\NN_x$ is learned, while the output mapping is fixed to identity. 

\paragraph{Physics-based model}
Special network structure could  be used to embed prior physical knowledge. For instance, let us consider a two degree-of-freedom mechanical system (\emph{e.g.}, a cart-pole system) with state $x=[x_1\; x_2\; x_3\; x_4]^\top$ consisting in two measured positions $x_1$, $x_3$ and two corresponding unmeasured velocities $x_2$, $x_4$, driven by an external force $u$. A physics-based model for this system is
\begin{equation}
\label{eq:pendulum}
 \begin{bmatrix}
    \dot x_1   \\ 
    \dot x_2 \\
    \dot x_3 \\
    \dot x_4
  \end{bmatrix} = 
  \begin{bmatrix}
   x_2 \\
   \NN_1(x, u; \theta) \\
   x_4 \\
   \NN_2(x, u;\theta)
  \end{bmatrix}, \qquad
	\ymodel = \begin{bmatrix}
			1 & 0 & 0 & 0 \\
			0 & 0 & 1 & 0
			\end{bmatrix}
			x,
\end{equation}
where the integral dynamics for positions are fixed in the parametrization, while the velocity dynamics are modeled by neural networks, possibly sharing some of their innermost layers. 
For discrete-time identification, \eqref{eq:pendulum} could be discretized through the numerical scheme of choice.

\subsection{Input/output structures}
When limited or no system knowledge is available, the following \emph{input/output} (IO) model structure may be used: 
\begin{equation}
\label{eq:IO_model_structure}
 \ymodel_{k} = \NN_{\rm IO}(\ymodel_{k-1}, \ymodel_{k-2}\dots, \ymodel_{k-n_a},
 u_{k-1},  u_{k-2}, \dots, u_{k-n_b}; \theta), 
\end{equation}
where $n_b$ and $n_a$ denote the input and output lags, respectively, and $\NN_{\rm IO}$ is a neural network of compatible size. For an IO model, the state $x_k$ can be defined in terms  of inputs and (noise-free) outputs at past time steps, \emph{i.e.}, 
\begin{equation}
\label{eq:IO_regressor_formula}
x_k = \left\{\ymodel_{k-1},  \ymodel_{k-2}, \dots, \ymodel_{k-n_a}, u_{k-1}, u_{k-2}, \dots, u_{k-n_b} \right\}.
\end{equation}
This state evolves simply by shifting previous inputs and outputs over time, and appending the latest samples, namely: 
\begin{equation}
\begin{split}
x_{k+1}\!=\!\overbrace{\left\{\ymodel_{k}, \ymodel_{k-1}, \dots, \ymodel_{k-n_a+1}, u_{k}, u_{k\!-\!1}, \dots, u_{k\!-\!n_b+1}\right\}}^{=f_{\rm IO}(x_k,\ymodel_k,u_k)},
\end{split}
\end{equation}
where the IO state update function $f_{\rm IO}(x_k,\ymodel_k,u_k)$ has been introduced for notational convenience.

The IO model structure only requires to specify the dynamical orders $n_a$ and $n_b$. If these values are not known a priori, they can be chosen through cross-validation.

\section{Training neural models}
\label{sec:training}

In this section, we present practical algorithms to fit the model structures  introduced in Section \ref{sec:model_structure} to the identification dataset $\Did$. 
For the sake of illustration, algorithms are detailed for  IO model structures~\eqref{eq:IO_model_structure}. 
The extension to state-space structures is then discussed in Subsection~\ref{sec:training_ss}. 

\subsection{Training I/O neural models}
\label{sec:training_IO}
The network parameters $\theta$ may be obtained by minimizing a cost function such as 
\begin{equation}
\label{eq:cost_function}
J(\theta) = \frac{1}{\nsamp} \sum_{k=0}^{N-1} \norm{\hat y_{k}(\theta) -   y_k}^2,
\end{equation}
where $\hat y_k$ is the model estimate at time $k$. 
For a dynamical model, different  estimates $\hat y_k$ can be considered in the cost~\eqref{eq:cost_function}, as discussed in the following paragraphs.

\paragraph{One-step prediction}
 The one-step  error loss $J^\mathrm{pred}(\theta)$ is constructed by plugging in \eqref{eq:cost_function} as estimate $\hat y_k$ the one-step prediction
$\hat y_{k}^{\mathrm{pred}} = \NN_{\rm IO}(\tilde x_{k}; \theta)$,
where $\tilde x_k$ is constructed as in~\eqref{eq:IO_regressor_formula}, but using \emph{measured} past outputs instead of the (unknown) noise-free outputs, \emph{i.e.}, $\tilde x_k = \{y_{k-1}$,  $y_{k-2}$, $\dots$, $y_{k-n_a}$, $u_{k-1}$, $u_{k-2}$, $\dots$, $u_{k-n_b} \}$.

The gradient of the cost function $J^\mathrm{pred}(\theta)$ with respect to $\theta$ can be readily obtained through back-propagation using modern deep learning frameworks.
This enables straightforward implementation of an iterative gradient-based optimization scheme to minimize $J^\mathrm{pred}(\theta)$.
The resulting one-step prediction error minimization algorithm can be  executed very efficiently on modern hardware since all time steps can be processed independently and thus in parallel, exploiting multiple CPU/GPU cores.  

For noise-free data,  one-step prediction error minimization   usually provides accurate results. Indeed, under the full model structure hypothesis, the minimum  of $J^{\mathrm{pred}}(\theta)$ is equal to $0$ and is achieved by all   parameters  $\theta \in \Theta_\So^{\Did}$.    
However, for noisy output observations,  the estimate $\hat y_{k}^{\mathrm{pred}}$   directly depends on  the noise affecting past outputs through the regressor $\tilde x_k$.   
The situation is reminiscent of the \emph{AutoRegressive with Exogenous input} (ARX)  linear predictor defined as
\begin{multline}
 \label{eq:ARX}
 \hat y_{k}^{\mathrm{ARX}} = a_1  y_{k-1} + a_2   y_{k-2} + \dots + a_{n_a}   y_{k-n_a}\\ 
 b_0 u_k+ b_1 u_{k-1}  + \dots b_{n_b} u_{k-n_b}, 
\end{multline}
and thoroughly studied in classic system identification~\citep{ljung:1999system}.
The minimizer of the ARX prediction error is generally \emph{biased}, unless very specific (and not particularly realistic) noise conditions are satisfied. Historically, the ARX predictor has been introduced for computational convenience---the resulting fitting problem can be solved indeed through linear least squares---rather than for its robustness to noise. 
In our numerical examples, we observed similar bias issues  when fitting neural model structures by minimizing $J^\mathrm{pred}(\theta)$ on noisy datasets.

\paragraph{Open-loop simulation}
In classic system identification for linear systems, the \emph{Output Error} (OE) predictor 
\begin{multline}
 \label{eq:OE}
 \hat y_{k}^{\mathrm{OE}} = a_1\hat y_{k-1}^{\mathrm{OE}} + a_2 \hat y_{k-2}^{\mathrm{OE}} + \dots + a_{n_a} \hat y_{k-n_a}^{\mathrm{OE}}\\ 
 + b_0 u_{k} + b_1 u_{k-1} + \dots b_{n_b} u_{k-n_b}, 
\end{multline}
defined recursively in terms of previous \emph{simulated} outputs provides an unbiased model estimate under the full model structure hypothesis, at the cost of a higher computational burden. In fact, minimizing the OE residual requires to solve a nonlinear optimization problem.
 
Inspired by these classic system identification results, in the neural modeling context 
we expect better noise robustness by minimizing the simulation error cost $J^{\mathrm{sim}}(\theta)$ obtained by using as estimate $\hat y_k$ in \eqref{eq:cost_function} the open-loop simulated output $y_k^{\mathrm{sim}}  = \NN_{\rm IO}(x_k^{\mathrm{sim}};\theta)$, with $x_{k}^{\mathrm{sim}}$ defined recursively in terms of previous \emph{simulated} outputs as
\begin{equation} 
		x_{k+1}^{\mathrm{sim}} = f_{\rm IO}(x_{k}^{\mathrm{sim}}, y_k^{\mathrm{sim}}, u_k) \;\; \text{for }k=0,\dots, \nsamp -1.  
\end{equation}

In principle, the cost function $J^{\mathrm{sim}}(\theta)$ and its gradient w.r.t. $\theta$ can be also computed using  a back-propagation algorithm, just as for $J^{\mathrm{pred}}(\theta)$.
However, from a computational perspective, simulating over time has an intrinsically sequential nature and offers scarce opportunity for parallelization. Furthermore, back-propagation through a temporal sequence, also known in the literature as \emph{Back-Propagation Through Time} (BPTT), has a computational cost that grows linearly with the sequence length $N$  \citep{williams1995gradient}.
In practice, as it will be illustrated in our numerical examples, minimizing the simulation error with a gradient-based method over the entire identification dataset $\Did$ may be inconvenient from a computational perspective.

\paragraph{Multi-step simulation}
A natural trade-off between full simulation and one-step  prediction is simulation over  \emph{subsequences} of the dataset $\Did$ with length $m<N$. 
The multi-step simulation error minimization algorithm presented here processes \emph{batches} containing $\batchsize$ subsequences extracted from $\Did$ in parallel to enable efficient implementation.

A batch is completely specified by a \emph{batch start vector} $s \in \mathbb{N}^{\batchsize}$ defining the initial time instant of each subsequence. Thus, for instance, the $j$-th output subsequence in a batch   
contains the measured output samples $\{y_{s_j},$ $y_{s_j+1},$ $\dots, y_{s_j+m-1}\}$ where $s_j$ is the $j$-th element of $s$.
For notational convenience, let us arrange the batch output subsequences in a three-dimensional \emph{tensor} 
$\tens {y}\in\mathbb{R}^{\batchsize \times \seqlen \times n_y}$ whose elements are
$\tens {y}_{j,t} = y_{s_{j}+t}$, 
 with batch index $j\!=\!0,1,\dots,\batchsize\!-\!1$ and time index  $t\!=\!0,1,\dots,\seqlen\!-\!1$. 
Similarly, let us arrange the batch input subsequences in a tensor $\tens {u}\in\mathbb{R}^{\batchsize \times \seqlen \times n_u}$ where
$\tens {u}_{j,t} = u_{s_{j}+t}.$

The $m$-step simulation  $\hat {\tens{y}}^{\mathrm{sim}}$ for all subsequences has the same tensor structure as $\tens{y}$ and is defined as 
\begin{subequations} \label{eq:minibatch_simulation_error}
\begin{align} 
    \hat {\tens{y}}_{j,t}^{\mathrm{sim}}  &= \NN( {\tens{x}}_{j,t}^{\mathrm{sim}},\theta)  
\intertext{where the regressor  $\tens{x}_{j,t}^{\mathrm{sim}}$ is recursively obtained  as} 
{\tens{x}}^{\mathrm{sim}}_{j,t+1} &= f_{\rm IO}({\tens{x}}^{\mathrm{sim}}_{j,t}, \hat {\tens{y}}_{j,t}^{\mathrm{sim}}, {\tens{u}}_{j,t}),  
\end{align}
\end{subequations}
for $t=0,\dots, \seqlen -1$.
The initial regressor $\tens{x}^{\mathrm{sim}}_{j,0}$ of each subsequence may be constructed by plugging past input and output measurements into \eqref{eq:IO_regressor_formula}, \emph{i.e.},
$\tens{x}^{\mathrm{sim}}_{j,0} = \{y_{s_j-1}$, $\dots$, $y_{s_j-n_a}$, $u_{s_j-1}$, $\dots$, $u_{s_j-n_b}\}$. 
In this way, the measurement noise enters in the $m$-step simulation  only at the initial time step $t=0$ of the subsequences, and therefore its effect is less severe than in the one-step prediction case.

A basic multi-step simulation error approach consists in minimizing the cost:
\begin{align}
J^{\rm{sim}}_{\batchsize,\seqlen}(\theta)&\!=\!\frac{1}{\batchsize \seqlen}\sum_{j=0}^{\batchsize-1}\sum_{t=0}^{\seqlen-1} \norm{\hat{\tens{y}}^{\rm sim}_{j,t} - {{\tens{y}}}_{j,t}}^2.   
\end{align}
Such an approach  outperforms one-step prediction error minimization in the presence of measurement noise.

In this paper, we further improve the basic multi-step simulation method by considering also the initial condition of the subsequences as free variables to be tuned, along with the network parameters $\theta$. 
Specifically, we introduce an optimization variable $\hidden{Y} = \{\hidden{y}_0,\hidden{y}_1,\dots,\hidden{y}_{\nsamp-1}\}$ with the same size and structure as $\Yid$.
The initial condition  $\tens{x}^{\mathrm{sim}}_{j,0}$ for the batch is constructed as
$\tens{x}^{\mathrm{sim}}_{j,0} = \{\hidden{y}_{s_j-1}, \dots, \hidden{y}_{s_j-n_a}, u_{s_j-1}, \dots, u_{s_j-n_b} \}$, with $j=0,\ldots,\batchsize-1$. 
By considering such an initial condition,  the measurement noise does not enter in the model simulation.   Thus, as in pure open loop simulation error minimization, bias issues are circumvented. 

Since we are estimating the initial conditions in addition to the neural network parameters, a price is paid in terms of an increased variance of the estimated model. In order to mitigate this effect, 
the variable  $\hidden{Y}$ used to construct the initial conditions $\tens{x}^{\mathrm{sim}}_{j,0}$ can be enforced to  represent the unknown, noise-free system output and thus to be consistent with~\eqref{eq:IO_model_structure}. To this aim, we introduce a regularization term   penalizing the distance between  $\hat {\tens{y}}^{\mathrm{sim}}$ and $\hidden{\tens{y}}$, where ${\hidden{\tens{y}}}$ is a tensor with the same structure as ${\tens{y}}$, but   containing samples from $\hidden{Y}$, \emph{i.e}, ${\hidden{\tens{y}}}_{j,t} = \hidden{y}_{s_{j}+t}$. 

Algorithm \ref{algo:minibatch} details   the steps required to train a dynamical neural model by multi-step simulation error minimization with initial state estimation. In Step 1, the neural network parameters $\theta$ and the ``hidden'' output variable $\hidden{Y}$ are initialized to (small) random numbers and to $\Yid$, respectively. Then, at each iteration $i=0,\ldots,n-1$ of the gradient-based training algorithm, the following steps are executed.   
Firstly, the batch start vector $s \in \mathbb{N}^\batchsize$ is selected with $s_j \in [\max({n_a, n_b}) \ \  \nsamp-\seqlen-1],\; j=0,1,\dots,\batchsize-1$ (Step 2.1). The indexes in $s$ may be either (pseudo)randomly generated, or chosen deterministically.\footnote{For an efficient use of the identification dataset $\Did$, $s$ has to be chosen in such a way that all samples are visited during training.}  
Then,  tensors ${{\tens{y}}}$, ${\hidden{\tens{y}}}$, ${{\tens{u}}}$, and $\tens{x}^{\mathrm{sim}}_{j,0}$ are populated with the corresponding samples in $\Did$ (Step 2.2). 
Subsequently, $m$-step model simulation is performed (Step 2.3) and the cost function $J_{\batchsize,\seqlen}(\theta,\hidden{Y})$ to be minimized is computed (Step 2.4). The cost $J_{\batchsize,\seqlen}(\theta,\hidden{Y})$ in~\eqref{eq:batch_cost} takes into account both the fitting criterion (thus, the distance between $\hat {\tens{y}}^{\mathrm{sim}}$ and $ {\tens{y}}$) and a regularization term penalizing the distance between $\hat {\tens{y}}^{\mathrm{sim}}$ and 
$\hidden{\tens{y}}$. Such a regularization terms aims at enforcing consistency of the hidden output $\hidden Y$ with the model structure~\eqref{eq:IO_model_structure}. A weighting constant     $\alpha \in (0 \ \ 1]$ balances the two objectives.  
Lastly, the gradients of the cost with respect to the optimization variables $\theta$, $\hidden{Y}$ are obtained through BPTT (Step 2.5) and the optimization variables are updated via gradient descent with \emph{learning rate} $\lambda$ (Step 2.6). Improved variants of  gradient descent  such as \emph{RMSprop} or \emph{Adam}~\citep{kingma2014adam} can be alternatively adopted at Step 2.6.

\begin{remark}
The computational cost of BPTT in $\seqlen$-step simulation is  proportional to $\seqlen$ (and not to $\nsamp$, which is the case for open-loop simulation). Furthermore, processing of the $\batchsize$ subsequences can be carried out independently, and thus in parallel on current hardware and software which support parallel computing. For these reasons, running multi-step simulation error minimization with $\seqlen \ll \nsamp$  is significantly faster than pure open-loop simulation error minimization.
\end{remark}

\begin{algorithm}
	\caption{Multi-step simulation error minimization}
	\label{algo:minibatch}
\small
	\textbf{Inputs}: identification dataset $\Did$; number of iterations $n$; batch size $\batchsize$; length of subsequences~$\seqlen$; learning rate $\lambda>0$;  weight $\alpha \in (0 \ \ 1]$.
\vspace*{-.0cm}\hrule\vspace*{.0cm}
	\begin{enumerate}[label=\arabic*., ref=\theenumi{}]
        \item  \textbf{initialize} the neural network parameters $\theta$ to a random vector   and the hidden output $\hidden{Y}$ to $\Yid$;
		\item  \textbf{for} $i=0,\ldots,\numiter-1$ \textbf{do}
		\begin{enumerate}[label=\theenumi{}.\arabic*., ref=\theenumi{}.\theenumii{}]
			\item \textbf{select} batch start indexes vector $s \in \mathbb{N}^q$;
			\item \textbf{define} tensors
			\begin{align*}
			& {{\tens{y}}}_{j,t}=y_{s_j+t}, \qquad
			 \hidden{{\tens{y}}}_{j,t}=\hidden{y}_{s_j+t}, \qquad
			 {{\tens{u}}}_{j,t}=u_{s_j+t}, \\
			 & \textrm{for  } j\!=\!0,1,\dots,\batchsize\!-\!1 \textrm{\ \ and\ \ } t\!=\!0,1,\dots,\seqlen\!-\!1
			\end{align*}
			and
			\begin{align*}
			& \tens{x}^{\mathrm{sim}}_{j,0} = \{\hidden{y}_{s_j-1}, \dots,   \hidden{y}_{s_j-n_a}, u_{s_j-1}, \dots, u_{s_j-n_b} \}, \\
			& \textrm{for  } j\!=\!0,1,\dots,\batchsize\!-\!1;
			\end{align*}
			\item \textbf{simulate} $\hat{\tens{y}}^{\rm sim}$ according to 
			\begin{align*} 
					\hat {\tens{y}}_{j,t}^{\mathrm{sim}}  &= \NN_{\rm IO}( {\tens{x}}_{j,t}^{\mathrm{sim}},\theta) \\
					{\tens{x}}^{\mathrm{sim}}_{j,t+1} &= f_{\rm IO}({\tens{x}}^{\mathrm{sim}}_{j,t}, \hat {\tens{y}}_{j,t}^{\mathrm{sim}}, {\tens{u}}_{j,t}) \\
					\! \textrm{for  } & j\!=\!0,1,\dots,\batchsize\!-\!1 \textrm{\ \ and\ \ } t\!=\!0,1,\dots,\seqlen\!-\!1; 
			\end{align*} 
			
			\item \textbf{compute} the cost 
			\begin{equation}
			\begin{split}
			\label{eq:batch_cost}
			J_{\batchsize,\seqlen}(\theta,\hidden{Y})&\!=\!\frac{\alpha}{\batchsize \seqlen}\sum_{j=0}^{\batchsize-1}\sum_{t=0}^{\seqlen-1} \norm{\hat{\tens{y}}^{\rm sim}_{j,t} - {{\tens{y}}}_{j,t}}^2  + \\
			        &+\frac{1\!-\!\alpha}{\batchsize \seqlen} \sum_{j=0}^{\batchsize-1}\sum_{t=0}^{\seqlen-1}  \norm{\hat{\tens{y}}^{\rm sim}_{j,t} - \hidden{{\tens{y}}}_{j,t}}^2;
			\end{split}
			\end{equation}

\item \textbf{evaluate} the gradients $\nabla_\theta J_{\batchsize,\seqlen}=\frac{\partial J_{\batchsize,\seqlen}}{\partial \theta}$ and  
			$\nabla_{\hidden{Y}} J_{\batchsize,\seqlen}=\frac{\partial J_{\batchsize,\seqlen}}{\partial \hidden{Y}}$ at the current values of $\theta$ and $\hidden{Y}$;
			\item  \textbf{update} optimization variables $\theta$ and $\hidden{Y}$:
			\begin{equation}
			\label{eq:SGD}
			\begin{split}
			 \theta &\leftarrow \theta - \lambda \nabla_\theta J_{\batchsize,\seqlen}  \\
			 \hidden{Y} & \leftarrow \hidden{Y} - \lambda \nabla_{\hidden{Y}} J_{\batchsize,\seqlen};
			\end{split}
			\end{equation}
		\end{enumerate}
	\end{enumerate}
	\vspace*{-.0cm}\hrule\vspace*{.1cm}
	\textbf{Output}:  neural network parameters $\theta$. 
\end{algorithm} 

\subsection{Training state-space neural models}
\label{sec:training_ss}
The fitting methods presented for the IO structure are applicable to the state-space structures introduced in Section~\ref{sec:ss_model_structure}, with the considerations   discussed below. 

\begin{itemize}	
\item For the fully observed state model structure~\eqref{eq:neural_ss_full}, adaptation of the one-step prediction error minimization method is straightforward. Indeed, the noisy measured state $y_k=x_k+\eta_k$ is directly used as regressor to construct the predictor, \emph{i.e.}, $\hat y_{k+1}^{\mathrm{pred}} = \NN_x(y_{k},u_k; \theta)$.
\item For model structures where the state is not fully observed, one-step prediction error minimization is not directly applicable as 
a one-step ahead prediction cannot be constructed in terms of the available measurements.
\item Simulation error minimization is directly applicable to state-space structures without special modifications, provided that it is feasible from a computational perspective.	
\item Algorithm~\ref{algo:minibatch} for multi-step simulation error minimization is also generally applicable for  state-space model structures.  Instead of the hidden output variable $\hidden{Y}$, a hidden state variable $\hidden{X}$ representing the (noise-free) state at each time step must be optimized along with the network parameters $\theta$ through gradient descent. However, if the state is not fully observed, $\hidden{X}$ cannot be initialized directly with measurements as was done in the IO case.
A convenient initialization of $\hidden{X}$ to be used in  gradient-based optimization can come from an initial state estimator, or exploiting physical knowledge. For instance, for the mechanical system in~\eqref{eq:pendulum}, a possible initialization for velocities is obtained through numerical differentiation  of the measured position outputs.
\end{itemize}	

\section{Numerical Example}
\label{sec:example}
The fitting algorithms for the model structures presented in this paper are tested  on a simulated nonlinear RLC circuit.  
 All computations are carried out on a PC equipped with an Intel i5-7300U 2.60~GHz processor and 32 GB of RAM.  
 The software implementation is based on the PyTorch Deep Learning Framework \citep{paszke:2017automatic}.  All the codes implementing the methodologies discussed in the paper and  required to reproduce the results are available on the on-line repository \url{https://github.com/forgi86/sysid-neural-structures-fitting}. Other examples concerning the identification  of a \emph{Continuously Stirred Tank Reactor} (CSTR) and a cart-pole system are available in the same repository.
\subsection{System description}
\begin{figure}
\centering
\begin{subfigure}{.4\textwidth}
  \centering
  \includegraphics[width=1.0\linewidth]{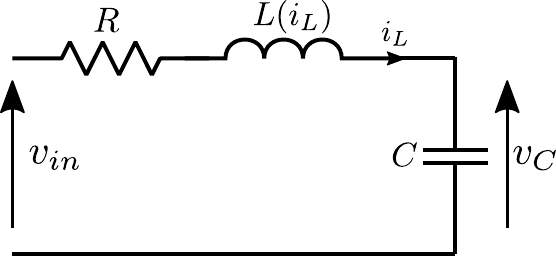}
  \label{fig:sub1}
\end{subfigure}%
\begin{subfigure}{.4\textwidth}
  \centering
  \includegraphics[width=.99\linewidth]{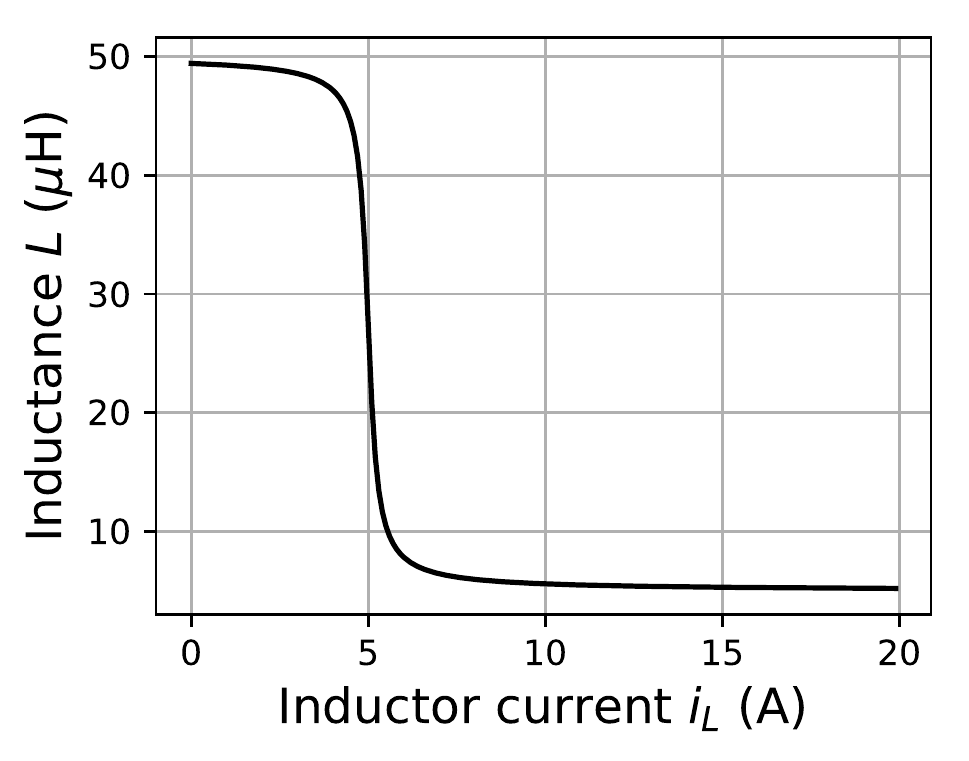}
  \label{fig:sub2}
\end{subfigure}
\caption{Nonlinear series RLC circuit used in the example (left) and nonlinear dependence of the inductance $L$ on the  inductor current $i_L$  (right).}
\label{fig:RLC}
\end{figure}

We consider the nonlinear RLC circuit in Fig.~\ref{fig:RLC} (left).
The circuit behavior is described by the continuous-time state-space equation 
\begin{equation}
\label{eq:RLC_sys}
\begin{bmatrix}
\dot v_C\\
\dot i_L
\end{bmatrix} = 
\begin{bmatrix}
  0           & \tfrac{1}{C}\\
 \tfrac{-1}{L(i_L)} & \tfrac{-R}{L(i_L)}\\
\end{bmatrix}
\begin{bmatrix}
v_C\\
i_L
\end{bmatrix} +
\begin{bmatrix}
0\\
\tfrac{1}{L(i_L)}
\end{bmatrix} 
v_{in},
\end{equation}
where $v_{in}~\rm(V)$ is the input voltage; $v_C~\rm(V)$ is the capacitor voltage; and $i_L~\rm(A)$ is the inductor current. The circuit parameters $R=3~\Omega$ and $C=270$~nF are fixed, while the inductance $L$ depends on $i_L$ as shown in Fig. \ref{fig:RLC} (right). Specifically, 
\begin{equation*}
 L(i_L) = L_0\bigg[\bigg(\frac{0.9}{\pi}\arctan\big(-\!5(|i_L|-5\big)+0.5\bigg) + 0.1 \bigg], 
\end{equation*}
with $L_0=50~{\mu \rm  H}$. 
The identification dataset $\Did$ is built by  discretizing~\eqref{eq:RLC_sys} using a $4$th-order Runge-Kutta method with a fixed step $T_s=0.5~\mu \text{s}$ and simulating the system for  $2~\text{ms}$.  $N=4000$ samples are  gathered. The input $v_{in}$ is filtered white noise with bandwidth $150~\text{kHz}$ and standard deviation $80~\text{V}$. An independent validation dataset is generated using as input $v_{in}$ filtered white noise   with  bandwidth $200~\text{kHz}$ and standard deviation $60~\text{V}$. 

The performance of the estimated models is assessed in terms of the $R^2$ index computed using the open-loop simulated model output.  
As  reference, a second-order linear OE model estimated on noise-free data using the \emph{System Identification Toolbox}  \citep{ljung:2015system} achieves an $R^2$ index of $0.96$ for $v_C$ and $0.77$ for $i_L$ on the identification dataset, and $0.94$ for $v_C$ and $0.76$ for $i_L$ on the validation dataset. 

 We consider for neural model structures the cases of ($i$) noise-free measurements of $v_C$ and $i_L$; ($ii$) noisy measurements of $v_C$ and $i_L$; ($iii$) noisy measurements of $v_C$ only. 

\subsection{Algorithm setup}
For gradient-based optimization, the Adam optimizer is used at Step 2.6 of Algorithm~\ref{algo:minibatch} to update the network parameters $\theta$ and the hidden variable $\hidden{Y}$ (or $\hidden{X}$ for state-space structures) used to construct the initial conditions $\tens{x}_{j,0}^{\mathrm{sim}}$. The learning rate $\lambda$ is adjusted  through a rough trial and error, with $\lambda$ taking values in the  range $[10^{-2}~10^{-6}]$, while the number of iterations $\numiter$ is chosen large enough to reach a plateau in the cost function. 
In Algorithm~\ref{algo:minibatch}, the weight $\alpha$ is always set to $0.5$.  
We tested different values for the sequence length $m$ and adjusted the batch size $q$ such that $qm \approx N$. 

 \begin{figure}
	\centering
	\includegraphics[width=.7\linewidth]{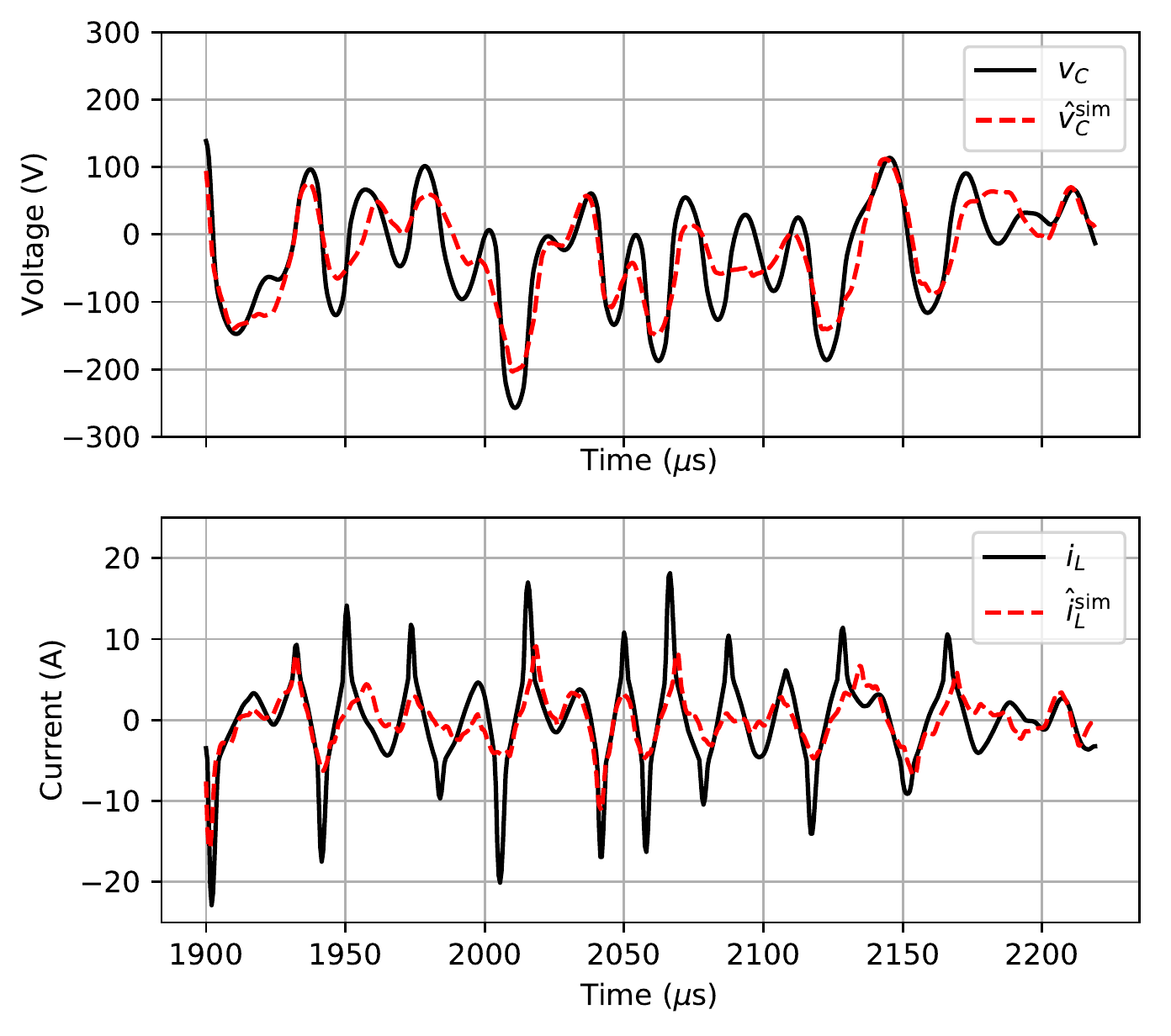}
	\caption{True output (black) and  simulated output (red) obtained by the state-space model trained using the one-step prediction error minimization approach in the presence of noise.}
	\label{fig:RLC_SS_val_1step_noise}
\end{figure}

\begin{figure}
\centering
   \includegraphics[width=.7\linewidth]{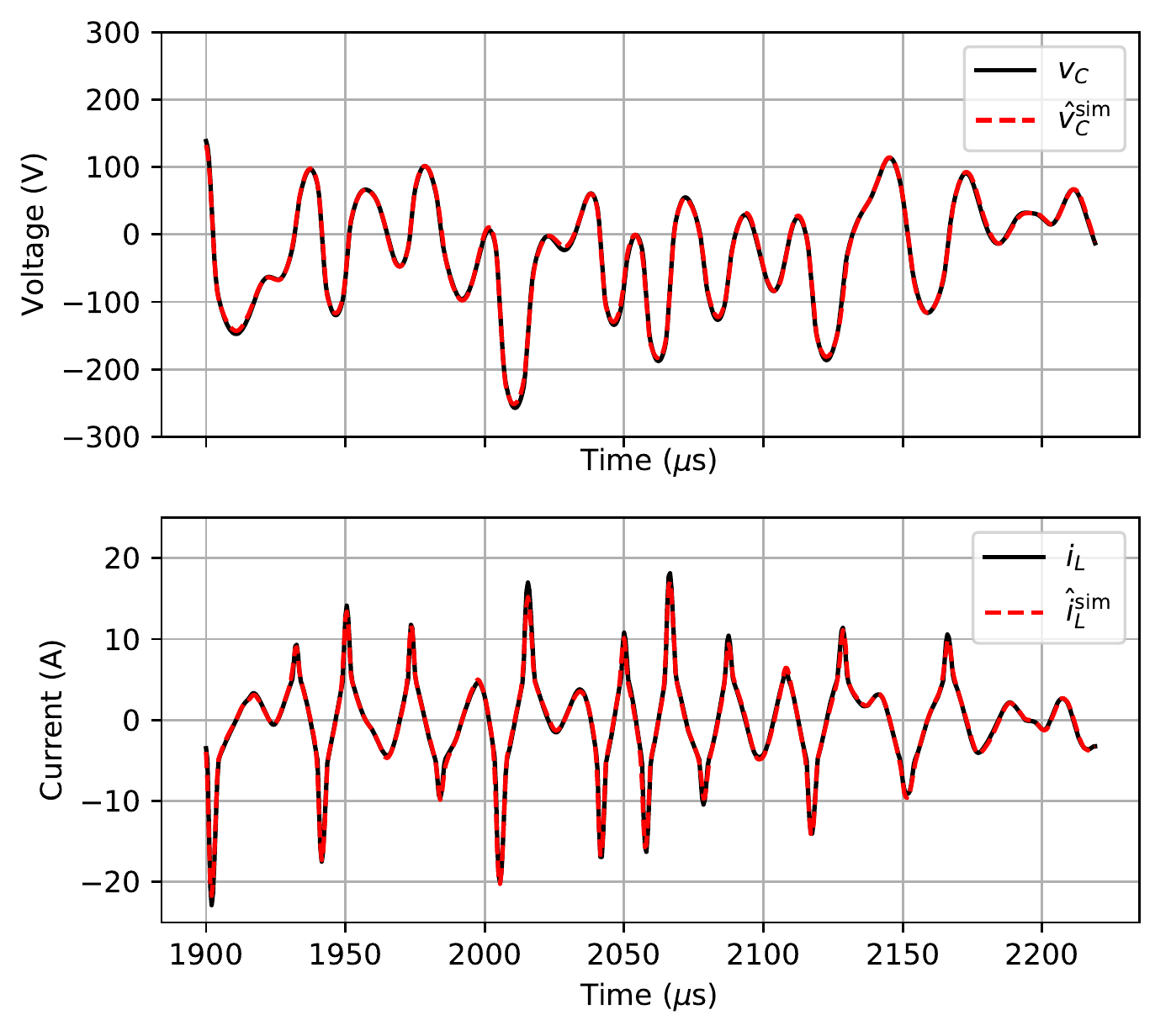}
\caption{True output (black) and  simulated output (red) obtained by the state-space model trained using the $64$-step simulation error minimization approach in the presence of noise.}
\label{fig:RLC_SS_val_64step_noise}
\end{figure}

\subsection{Results}
\emph{\bf{($i$) Noise-free measurements of $v_C$ and $i_L$}}
\ \\
Since in this case the system state is supposed to be measured, we use the fully-observed state model structure~\eqref{eq:neural_ss_full}.
The neural network $\NN_x$ modeling the state mapping has a sequential feedforward structure with three input units ($v_C$, $i_L$, and $v_{in}$); a hidden layer with 64 linear units followed by ReLU nonlinearity; and two linear output units---the two components of the state equation to be learned. 
Having a noise-free dataset, we expect good results from one-step prediction error minimization.
Thus, we fit the model using this approach over $\numiter=40000$ iterations with learning rate $\lambda= 10^{-4}$. The time required to train the network is 114 seconds.
The fitted model describes the system dynamics with high accuracy. On both the identification and the validation datasets, the model $R^2$ index in open-loop simulation is above $0.99$ for $v_C$ and $0.98$ for $i_L$.

\emph{\bf{($ii$) Noisy measurements of $v_C$ and $i_L$}}
\ \\
We consider the same identification problem above, with observations of $v_C$ and $i_L$ corrupted by an additive white Gaussian noise with standard deviation $10~\rm V$ and $1~\rm A$, respectively. This corresponds to a \emph{Signal-to-Noise Ratio} (SNR) of  20 dB and 13 dB on  $v_C$ and $i_L$, respectively.  
Results for the one-step prediction error minimization approach on the validation dataset are shown in Fig.~\ref{fig:RLC_SS_val_1step_noise}. 
 The  $R^2$ index in validation drops to $0.73$ for $v_C$ and $0.03$ for  $i_L$.  
It is evident that noise has a severe impact on the performance of the one-step method. 

In the presence of noise, better performance is expected from a multi-step simulation error minimization approach. Thus, we fit the same neural state-space model structure using Algorithm \ref{algo:minibatch} over $\numiter=15000$ iterations, with $\lambda=10^{-3}$, and  randomly extracted batches of $q=62$ subsequences, each of length $m=64$.    
The results are in line with expectations. Indeed, we recover similar performance as one-step prediction error minimization in the noise-free case ($R^2$ index of $0.99$ on $v_C$ and $0.98$ on $i_L$ on both the identification and the validation datasets). Time trajectories of the output  are reported in Fig.~\ref{fig:RLC_SS_val_64step_noise}. For the sake of visualization, only a portion of the validation dataset is shown.   
The total run time of Algorithm \ref{algo:minibatch} is 182 seconds---about 60\% more than the one-step prediction error minimization method. 

Open-loop simulation error minimization is also tested. This method yields the same performance of $64$-step simulation error minimization in terms of $R^2$ index of the fitted model. However, it takes about two hours to execute $\numiter=10000$ iterations required to reach a cost function plateau.

\emph{\bf{($iii$) Noisy measurements of $v_C$ only}}
\ \\
We consider the  case where only the voltage $v_{C}$ is measured and corrupted by an additive white Gaussian noise with standard deviation $10$~V.  The IO model structure in \eqref{eq:IO_model_structure} is used with $n_a=2$ and $n_b=2$. The neural network $\NN_{\rm IO}$ is characterized by four input units (corresponding to $n_a=2$ previous values of $v_C$ and $n_b=2$ previous values of $v_{in}$);  a  hidden layer with 64 linear units followed by  ReLU nonlinearity; and a linear output unit representing the output  value $v_C$. 
As in case ($ii$), the one-step prediction error minimization approach delivers  unsatisfactory results due to the presence of measurement noise. 
Thus, we fit the model using the multi-step method described in Algorithm \ref{algo:minibatch} over $\numiter=15000$ iterations, with $\lambda=10^{-3}$, $m=32$, and $q=124$.
The total runtime of the algorithm is 192 seconds.
The $R^2$ index of the fitted model is above  $0.99$ on both the identification and the validation dataset, thus even larger than the $R^2$ index achieved by the OE model estimated on noise-free data.


\section{Conclusions and follow-up}
\label{sec:conclusions}
In this paper, we have presented neural model structures and fitting criteria for the identification    of dynamical systems.  A custom method minimizing a regularized multi-step simulation error criterion has been  proposed and compared with one-step prediction error and simulation error minimization.  

The main strengths of the presented framework are its versatility to describe complex non-linear systems, thanks to the neural network flexibility; its robustness to the measurement noise, thanks to the multi-step simulation error criterion
with initial condition estimation; and the possibility to exploit parallel computing to train the network and optimize the initial conditions, thanks to the division of the dataset into small-size subsequences. 

Current and future research activities are devoted to: ($i$) the formulation of proper fitting criteria and optimization algorithms for direct learning of continuous-time systems and systems described by partial differential equations, without introducing numerical discretization; ($ii$) the development of computationally efficient algorithms for estimation and control based on the  neural dynamical models. 

\section*{Acknowledgements}
 The authors are grateful to Dr.~Giuseppe~Sorgioso for the fruitful discussions on the properties of the back-propagation through time algorithm.

\bibliography{ms}             
                                                   
\end{document}